\documentclass[manuscript,screen]{acmart}

\usepackage{caption}
\usepackage{natbib}
\usepackage{amsmath}   
\usepackage{mathtools}
\usepackage{soul}      
\usepackage{hyperref}
\usepackage{algorithm}
\usepackage[noend]{algpseudocode}
\usepackage{multirow}

\AtBeginDocument{%
  }

\setcopyright{none}
\renewcommand\footnotetextcopyrightpermission[1]{}

\renewcommand{\shortauthors}{Saini et al.}

\makeatletter
\fancypagestyle{standardpagestyle}{%
  \fancyhf{}%
  \fancyhead[LE,RO]{\ACM@pagestyle@head}%
  \fancyhead[LO]{\shortauthors}%
  \fancyhead[RE]{\@shorttitle}%
  \renewcommand{\headrulewidth}{\z@}%
  \renewcommand{\footrulewidth}{\z@}%
}
\makeatother

\begin{document}

\sloppy

\title{Evolved Sample Weights for Bias Mitigation: Effectiveness Depends on the Fairness Objective}

\author{Anil K. Saini}
\email{anil.saini@cshs.org}
\orcid{0000-0003-1819-9129}
\affiliation{%
  \institution{Cedars-Sinai Medical Center, Los Angeles}
  \state{California}
  \country{USA}}

\author{Jose Guadalupe Hernandez}
\orcid{0000-0002-1298-5551}
\affiliation{%
  \institution{Cedars-Sinai Medical Center, Los Angeles}
  \state{California}
  \country{USA}}
\email{jose.hernandez8@cshs.org}

\author{Emily F. Wong}
\orcid{0000-0003-1819-9129}
\affiliation{%
  \institution{Cedars-Sinai Medical Center, Los Angeles}
  \state{California}
  \country{USA}}
\email{emily.wong@cshs.org}

\author{Debanshi Misra}
\affiliation{%
  \institution{University of California, Los Angeles}
  \state{California}
  \country{USA}}
\email{debanshi@ucla.edu}

\author{Tiffani J. Bright}
\orcid{0000-0002-0894-0107}
\affiliation{
    \institution{Cedars-Sinai Medical Center, Los Angeles}
    \state{California}
    \country{USA}
}
\email{tiffani.bright@cshs.org}

\author{Jason H. Moore}
\orcid{0000-0002-5015-1099}
\affiliation{%
  \institution{Cedars-Sinai Medical Center, Los Angeles}
  \state{California}
  \country{USA}}
\email{jason.moore@csmc.edu}

\renewcommand{\shortauthors}{Saini et al.}


\begin{abstract}
Machine learning models trained on real-world data may inadvertently make biased predictions that negatively impact marginalized communities.
Reweighting, which assigns a weight to each data point used during model training, can mitigate such bias, though sometimes at the cost of predictive accuracy.
In this paper, we investigated this trade-off by comparing three methods for generating these weights: (1) evolving them using a Genetic Algorithm (GA), (2) computing them using only dataset characteristics, and (3) assigning equal weights to all data points.
Model performance under each strategy was evaluated using paired predictive and fairness metrics.
We used two predictive metrics (accuracy and area under the Receiver Operating Characteristic curve) and two fairness metrics (demographic parity and subgroup false negative fairness).
By conducting experiments on eleven publicly available datasets (including two medical datasets), we show that evolved sample weights can produce models that achieve better trade-offs between fairness and predictive performance than alternative weighting methods. 
However, the magnitude of these benefits depends strongly on the choice of fairness objective. 
Our experiments reveal that the evolved weights were most effective when optimizing for demographic parity---independent of choice of the performance objective---yielding better performance than other weighting strategies on the largest number of datasets.
\end{abstract}



\maketitle
\thispagestyle{plain}
\pagestyle{plain}
\fancyfoot{}

\section{Introduction}
While machine learning (ML) has revolutionized numerous industries, it has also demonstrated the ability to perpetuate racial, gender, and other biases embedded within a dataset~\cite{barocas2023fairness}. 
In areas where these ML systems are used to make high-stakes decisions, such as healthcare, algorithmic bias can have unintended negative consequences (e.g., widening health disparities).
All ML, regardless of whether the models are learning through supervised, unsupervised, or semi-supervised approaches, requires data.
Given that the biases embedded within the data may be unknown, there is a risk of algorithmic bias for all ML methods. 
Bias can arise from various sources, such as the use of incorrect features and the lack of diversity during sampling \cite{mehrabi2021survey}.
Bias may also be introduced by the configuration of an algorithm (e.g., the choice of optimization functions or regularization) \cite{mehrabi2021survey}.

Bias can be ameliorated at various stages of employing ML models:
pre-processing, which modifies data prior to training and evaluation;
in-processing, which involves tuning the algorithm during the training process; 
and post-processing, which adjusts predictions after training. 
Reweighting is a widely used pre-processing approach to mitigate bias in model predictions. 
It involves assigning weights to data points in the training set (called `sample weights' hereafter) that are utilized by ML models during training to adjust the contribution of different data points to the loss function of the model. 
While loss functions vary across ML models, they all quantify the difference between predicted and true values of the target variable to guide the model's internal optimization.
For example, neural networks typically use cross-entropy loss during gradient descent.

Two prominent strategies exist for computing sample weights. The first is a deterministic approach, where weights are derived directly from dataset characteristics (e.g., Kamiran \& Calders~\cite{kamiran2012data}).
The second is an optimization-based approach, where weights are evolved through an evolutionary algorithm to jointly improve predictive performance and fairness with respect to a specific machine learning model (e.g., La Cava~\cite{la2023optimizing}, and Hoitsma et al.~\cite{hoitsma2025mitigating}).
Deterministic reweighting has been shown to improve model fairness~\cite{Park,wong2024}, although sometimes at the cost of accuracy~\cite{blow2024comprehensive}.  
Recent work suggests that evolutionary approaches to learning sample weights (e.g., La Cava~\cite{la2023optimizing}) can more effectively optimize predictive error and fairness simultaneously than other non-reweighting methods (e.g., GerryFair~\cite{kearns2018preventing}).
However, the relative performance of evolved weights versus deterministic weights in terms of the trade-off between fairness and accuracy, under different combinations of predictive (e.g., accuracy) and fairness (e.g., demographic parity difference) metrics, has not been systematically examined.

In this work, we address this gap by making the following contributions:
\begin{enumerate}
    \item We compare three methods for computing sample weights: (i) evolving them using a Genetic Algorithm, (ii) deriving them from dataset characteristics, and (iii) assigning equal weights.
    \item We evaluate these reweighting strategies under multiple combinations of predictive and fairness metrics.
\end{enumerate}

Across eleven publicly available datasets, our experiments show that evolved sample weights yield models with minimal trade-offs between predictive performance and fairness compared to alternative weighting methods. 
Importantly, the extent of these improvements depends on the specific fairness metric used during optimization.





\section{Background and Related Work}

\begin{figure}[!ht]
\centering
\includegraphics[width=0.7\textwidth]{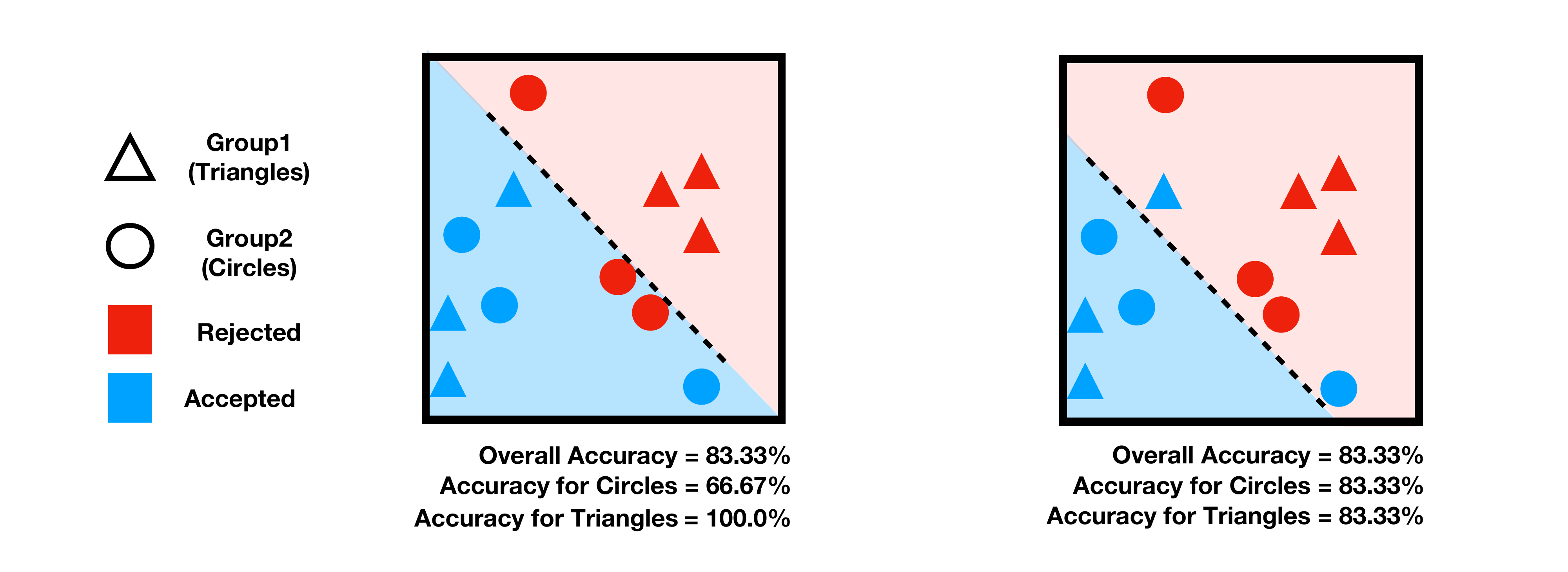}
\caption{An example of two different decision boundaries producing the same overall accuracy during training on a given dataset. The decision boundary on the left makes less fair predictions than the decision boundary on the right.}
\label{fig:toy_example}
\end{figure}

Algorithmic bias refers to systematic errors in the modeling process that produce lower-quality or less desirable predictions for certain historically disadvantaged communities, such as people of color and women~\cite{kordzadeh2022algorithmic}. Beyond data quality and variable selection, algorithmic bias can also result from a particular model's decision boundary.
To illustrate this point, consider a case study on hiring decisions.
Employers typically prefer to streamline administrative processes, such as hiring, especially when they receive a large volume of applications.
In this scenario, it would be useful to build an ML model that predicts the success of prospective candidates. 
An ML model trained on previous hiring decisions may have many potential decision boundaries.
Figure~\ref{fig:toy_example} displays two illustrative decision boundaries.
Applicant groups are represented by circles and triangles, and each point’s true label (accepted or rejected) is indicated by its solid fill color.
For both boundaries, the area above the diagonal (shaded in red) denotes a rejection prediction, and the area below (shaded in blue) denotes an acceptance prediction.

While both decision boundaries yield the same overall accuracy, their predictive performance varies between the two groups: the accuracy is higher for Triangles than Circles for the leftmost boundary in Figure \ref{fig:toy_example}, whereas the accuracy is the same for both groups with the rightmost boundary.
This example illustrates how modeling decisions, such as selecting decision boundaries, can negatively bias predictions against certain groups, while also showing how modeling decisions can reduce disparities without sacrificing overall predictive performance.

\subsection{Measuring Bias}


Multiple metrics exist to quantify the fairness of predictions made by an ML model \cite{barocas2023fairness}. 
Each metric is specific to the application context and attempts to quantify unique properties (false negative rate, accuracy, etc.) of the predictions for individuals belonging to different groups.
Generally, since the fairness metrics try to measure the disparity in the model predictions between groups, lower values on these metrics indicate less disparity and hence greater fairness.
From the available metrics, we choose two metrics, one that does not consider the types of incorrect predictions (demographic parity) and one that does take such information into account (subgroup false negative fairness).

Before describing these metrics, we first must specify \textit{sensitive} and \textit{non-sensitive} features in the dataset under consideration.
`Sensitive' (or `protected') features specify the attributes of individuals a model should be sensitive to when assessing group fairness (e.g., race, gender), whereas all other features are considered non-sensitive.
Based on the values of sensitive attributes, each data point can fall into one of the groups defined by a combination of those sensitive attributes. 
For example, `Black women younger than 25' would be one of the groups when the sensitive attributes are race, gender, and age.
Let $G\in \mathcal{G}$ be one such group.
For some metrics, in a binary classification setting, one value of the target variable is considered `favorable' (or desired), while the other value is not.
This favorable value is usually represented by 1.
For example, in a dataset containing records of applicants applying for loans, having received an approval would be considered favorable (and assigned a value of 1).

\textbf{Demographic Parity (DP)~\cite{weerts2023fairlearn}:} This metric measures the difference between the largest and the smallest group-level `acceptance rate', which is defined as the proportion of individuals belonging to the group receiving a favorable prediction. 
DP can be given mathematically as
$$DP = \max_{G_1,G_2}[Pr(\hat{Y}=1|X\in G_1)- Pr(\hat{Y}=1|X\in G_2)],$$
where $\hat{Y}$ is the predicted value of the target variable, $G_1, G_2$ are two subgroups withing the dataset, $X$ is a given data point in the dataset, and $Pr(.)$ is the probability estimate.

\textbf{Subgroup False Negative Fairness (SFN)~\cite{la2023optimizing,kearns2018preventing}:} It captures the maximum deviation of a model’s performance in terms of false negative rate among any one group in $\mathcal{G}$. The deviation is normalized by the probability of observing an individual from that group having positive labels.
The SFN can be given mathematically by
$$SFN = \max_{G\in \mathcal{G}}[\alpha(G)\beta(G)].$$ 
The term $\alpha$ denotes the probability of getting positive labels in the group $G$, and the term $\beta$ refers to the absolute difference between the overall false negative rate and the false negative rate within the group $G$.

Note that for many real-world applications (including medical diagnosis), a missed detection can be more costly than a false alarm. 
Therefore, we used the Subgroup False Negative Fairness as a proof of concept in our experiments.
Due to computational constraints, we did not use Subgroup False Positive Fairness, although it can be computed in a similar manner to SFN.

\subsection{Mitigating Bias}
Many techniques have been proposed to make ML algorithms less biased~\cite{chen2023algorithmic}.
Most of these methods can fall into three categories:
\begin{itemize}
    \item \textbf{Pre-processing:} Methods in this category either change some properties of the data, modify some values in the data, or change the loss function contribution of data points used by the model. 
    Examples include assigning weights to data points in the training data to be used by the loss function~\cite{kamiran2012data,jiang2020identifying}, modifying labels for some data points in the dataset~\cite{kamiran2012data}, feature selection~\cite{xing2021fairness}, and feature transformation~\cite{calmon2017optimized}. 
    
    \item \textbf{In-processing:} These methods modify the ML models directly to reduce bias in their predictions. 
    For example, in Zhang et al. \cite{zhang2018mitigating}, the authors add an adversarial model that predicts the sensitive attributes for a given predicted output. 
    The combination of the original predictor model and the adversarial model gives rise to a model with better values on fairness metrics. 
    
    \item \textbf{Post-processing:} These methods adjust model predictions to make them less discriminatory. 
    Examples include methods that change the threshold for risk scores of an already existing, possibly discriminatory, predictor \cite{hardt2016equality}.
\end{itemize}

\subsection{Reweighting}
\label{sec:reweigh}
In this work, we focus on the Reweighting method \cite{kamiran2012data}, which mitigates bias by assigning weights to data points in the training dataset. These weights are used while computing the loss function during the training phase and hence adjust each data point's contribution to the training loss. In this work, we will focus on the classifiers that support the use of weights in this way.

In its most basic form, through reweighting, we assign higher weights ($>$ 1) to data points that possess one of the following two characteristics:
(1) individuals belonging to underrepresented groups that possess desirable values of the target variable (e.g., female candidates getting accepted for a STEM job), 
and (2) individuals belonging to overrepresented groups that possess undesirable values of the target variable (e.g., male candidates getting rejected for a STEM job) \cite{kamiran2012data}.
All other data points are assigned lower weights ($<$ 1).
The weight assigned for a given data point $X$ is defined by:
\begin{equation}
    W(X) = \frac{P(S=X(S))\cdot P(Class = X(Class))}{P(S=X(S)\wedge Class = X(Class))},
\end{equation}
where $X(S)$ is the value of the sensitive feature in $X$, $X(Class)$ is the value of the target variable in $X$, $P(S=X(S))$ is the probability of observing a measurement in the dataset with the value of sensitive feature being $X(S)$, $P(Class = X(Class))$ is the probability of a given observation having value of the target variable as $X(Class)$, and  $P(S=X(S) \wedge Class=X(Class))$ is the probability of a given observation having the value of sensitive feature as $X(S)$ and target variable as $X(Class)$.

\begin{table}[h]
\centering
\caption{Example dataset to illustrate the Reweighting Method. The column titled `Weight' is calculated using the Reweighting procedure.} 
{\begin{tabular}{lllllll}
\toprule
Race & Position   & Oral  & Written & Combined & Promotion & Weight \\
\midrule
W    & Captain    & 89.52 & 95      & 92.808   & 1         & 0.84   \\
W    & Captain    & 80    & 95      & 89       & 1         & 0.84   \\
W    & Captain    & 82.38 & 87      & 85.152   & 1         & 0.84   \\
W    & Captain    & 88.57 & 76      & 81.028   & 0         & 1.4    \\
H    & Lieutenant & 76.19 & 84      & 80.876   & 0         & 0.6    \\
H    & Captain    & 76.19 & 82      & 79.676   & 0         & 0.6    \\
W    & Captain    & 76.19 & 82      & 79.676   & 1         & 0.84   \\
H    & Lieutenant & 70    & 84      & 78.4     & 1         & 1.8    \\
W    & Captain    & 73.81 & 81      & 78.124   & 0         & 1.4    \\
W    & Lieutenant & 84.29 & 72      & 76.916   & 1         & 0.84  \\
\bottomrule
\end{tabular}}
\label{tab:reweigh_ex}
\end{table}

Reweighting can be illustrated with the following example. 
Table \ref{tab:reweigh_ex} shows a dataset containing the oral, written, and combined test scores for a promotion exam for a Fire Department (adapted from the RICCI dataset \cite{miao2010did}). 
The race and current position of each test taker are also given. 



    


    

\begin{table}[]
\centering
\caption{Intermediate and final values calculated during the reweighting procedure for the example given in Section \ref{sec:reweigh}.} 
{\begin{tabular}{cccccc}
\toprule
Race (S) & Promotion (C) & $P(S=X(S))$                       & $P(C = X(C))$             & $P(S=X(S) \wedge C = X(C))$ & $W(X)$ \\
\midrule
W    & 1         & 7 / 10 & 6/10 & 5/10                 & 0.84 \\
W    & 0         & 7 / 10 & 4/10 & 2/10                 & 1.4  \\ 
H    & 1         & 3 / 10 & 6/10 & 1/10                 & 1.8  \\
H    & 0         & 3 / 10 & 4/10 & 2/10                 & 0.6 \\ 
\bottomrule
\end{tabular}}
\label{tab:dw2}
\end{table}

    

    
Using `Race’ as the sensitive feature and `Promotion’ as the target variable yields four combinations of Race and Promotion.
For each combination, we can calculate the weights according to Equation 1 (see Table \ref{tab:dw2}).
The resulting weights for each data point in Table \ref{tab:reweigh_ex} can be found in the last column. 
Note that even though weights are assigned to all points in the training dataset, only $2^{(|sensitive\_attributes| + 1)}$ weights (combination of different values for sensitive features and the target variable) are computed for binary classification datasets with binary sensitive features.

We refer to the weighting strategy given by Kamiran \& Calders~\cite{kamiran2012data} as `Deterministic Reweighting' in order to differentiate it from `Evolved Reweighting' described in the next section.

\section{Genetic algorithm to optimize sample weights}
Genetic Algorithms (GAs) \cite{mitchell1998introduction} are a collection of methods that draw inspiration from the theory of natural selection to solve optimization problems by initializing a set of potential solutions (i.e., population) and evolving those solutions to optimize one or more objective functions.  
GAs have been successfully used in prior research for multi-objective optimization~\cite{coello2000updated}.

In this work, we describe a GA for discovering a set of sample weights optimized for both predictive performance (e.g., AUROC) and fairness (e.g., demographic parity); the predictive metric is maximized while the fairness metric is minimized.
The sample weights evolved using the GA are used during training to mitigate bias.
The GA requires the following inputs: population size, weight dimensionality, maximum number of generations, a machine learning method, and training data.
The dimensionality of weights ($ind\_size$) is determined by the total possible combination of values from sensitive attributes and the target variable; all other inputs are user-specified.

\begin{algorithm}[h]
\caption{Genetic Algorithm for Reweighting}
\label{alg:rga}
\begin{algorithmic}[1]
\Procedure{Evaluate}{$pop, ml, X, y$}
\For{ind in pop}
    \State assigned\_weight = Expand\_and\_Assign(ind) \Comment{Assign weights for all data points}
    \State ind[`pred'] = Pred(ml, X, y, assigned\_weight)
    \Comment{Predictive metric (e.g., AUROC) on validation set}
    \State ind[`fair'] = Fair(ml, X, y, assigned\_weight)
    \Comment{Fairness metric (e.g. DP) on validation set}
\EndFor
\EndProcedure
\Procedure{ReweightingGA}{$pop\_size, ind\_size, max\_gen, ml, X, y$}
\State pop = InitializePop($pop\_size$, $ind\_size$)
\Comment{Generate $pop\_size$ weight vectors of length $ind\_size$}
\State Evaluate(pop, ml, $X$, $y$)

\State all\_evaluated = pop
\Comment{Archive of all evaluated individuals}

\For{$g$ in $\{1,\ldots,max\_gen\}$}
    \State AssignRankAndCrowding(pop)
    \Comment{Non-dominated sorting + crowding distance}

    \State offspring = [ ]
    \While{$|\text{offspring}| < pop\_size$}
        \State parent\_a = NonDomBinaryTournament(pop)
        \State parent\_b = NonDomBinaryTournament(pop)

        \State child = Crossover(parent\_a, parent\_b)
        \State child = Mutation(child)
        \State offspring.append(child)
    \EndWhile

    \State Evaluate(offspring, ml, $X$, $y$)
    \State all\_evaluated = all\_evaluated + offspring

    \State combined $\leftarrow$ pop + offspring
    \State AssignRankAndCrowding(combined)
    \State pop $\leftarrow$ SurvivalSelection(combined, $pop\_size$)
    \Comment{Fill by rank; break ties by crowding distance}
\EndFor

\State \textbf{Return} all\_evaluated
\Comment{All evaluated sample weights}
\EndProcedure

\end{algorithmic}
\end{algorithm}


Algorithm \ref{alg:rga} outlines our GA based on NSGA-II~\cite{deb2002fast}, which we describe further in this section.
At the beginning of an evolutionary run, the starting population is initialized with $pop\_size$ sample weights of size $ind\_size$; values for these weights are drawn at random from a uniform distribution between $[0.0,2.0)$.
Each sample weight's length corresponds to the number of unique combinations for sensitive features and the target variables: $2^{(|sensitive\_attributes| + 1)}$.
Once the initial population is constructed, each set of weights and training data is used to train a machine learning model.
When a sample weight is evaluated, we assign the appropriate weight to all the data points in the training dataset that correspond to the specific combination of sensitive features and target variable (line 3 in Algorithm \ref{alg:rga}). 
This weight assignment is similar to how weights are calculated in Table \ref{tab:dw2} and assigned to data points in Table \ref{tab:reweigh_ex} in the example from Section \ref{sec:reweigh}. 


We apply 10-fold cross-validation to compute predictive performance (e.g., AUROC) and fairness (e.g, DP) scores on the training data. 
These scores (i.e, the average predictive performance and the average fairness across 10 folds) determine each individual’s Pareto front rank and crowding distance (line 11 in Alg.~\ref{alg:rga}; see Deb et al.~\cite{deb2002fast}, Section III-B). An individual’s rank reflects its Pareto optimality relative to others: the first (best) front consists of nondominated solutions, where solution X dominates solution Y if X is at least as good as Y in all objectives and strictly better in at least one. 
Subsequent fronts are formed iteratively from the remaining solutions.
The crowding distance measures how isolated an individual is within its front; higher distances are preferred as they help preserve diversity. 
These two attributes---Pareto front rank and crowding distance---guide parent selection (line 14-15 in Alg.~\ref{alg:rga}) using Nondominated Binary Tournament Selection. 
In each selection event, two individuals are randomly selected from the population and then compared: those with higher ranks are discarded first, then among those remaining, individuals with lower crowding distance are removed.
If multiple candidates remain, one is randomly chosen. 

Two parents are needed to generate a single offspring (i.e., parent selection is used twice per offspring created).
There is a $0.8$ probability we use crossover to generate a single offspring from both parents, where each value of the offspring's weights comes from either parent with equal probability. Otherwise, the first parent is directly returned as the offspring. 
Once the offspring is constructed, there is a $0.1$ probability of a point mutation applied to individual values within the set of weights; the magnitude of a point mutation comes from a normal distribution with a mean of $0.0$ and a standard deviation of $1.0$. The value of each element is capped on both sides to be in the $[0.0, 2.0]$ range.
A total of $pop\_size$ offspring are constructed and then set as the new population.

After generating offspring, we evaluate them like their parents (line 19 in Alg. \ref{alg:rga}) and assign Pareto front ranks and crowding distances relative to the combined set of parents and offspring (line 22). 
Survival selection (line 23) then reduces this combined set back to the original population size by retaining Pareto-optimal individuals. 
The individuals from the first (best) front are selected first; if this front contains too many, those with higher crowding distance are prioritized.
If more individuals are needed, pipelines from subsequent fronts are added in order of front rank, again preferring those with higher crowding distance.
The resulting survivors form the next generation, and the same evolutionary cycle of evaluation, selection, and reproduction is repeated for a specified number of generations.
The values we use for crossover and mutation probabilities are typical for the GA literature; prior GA works suggest a low mutation probability ($<=$ 0.1) and a high crossover probability ($>=$0.5)~\cite{hassanat2019choosing}.


After the final generation, a Pareto front is constructed from all sample weights evaluated throughout the evolutionary search as follows.
After retrieving the evaluated sample weights, we retrain the models on the full training set and construct Pareto fronts based on their performance and fairness scores on the test set.

The method to evolve weights described here is similar to the ones used by La Cava~\cite{la2023optimizing} and Hoitsma et al.~\cite{hoitsma2025mitigating}, with one important distinction: we do not directly optimize the weight vector for the whole dataset, but instead a weight vector whose length corresponds to the combination of possible values of the sensitive features and the target variable, which is typically less than the total number of samples in the training dataset.

\section{Methods}
\label{sec:methods}

\subsection{Comparing Reweighting Methods}
\label{sec:exp_conds}
We conducted experiments to evaluate the three reweighting methods across four combinations of predictive and fairness metrics.
As discussed in the following sections, each method is assessed based on its ability to jointly optimize both the predictive and fairness metrics (or objectives).
We used the following three methods for calculating sample weights: (1) equal weights, (2) deterministic weights~\cite{kamiran2012data}, and (3) evolved weights.
To measure predictive performance, we used two commonly used metrics~\cite{rainio2024evaluation}: (a) accuracy (ACC), which focuses on the number of accurate predictions without considering the kind of errors (false negatives or false positives), and (b) Area Under the Receiver Operating Characteristic curve (ROC), which takes into account the true positive and false positive rates.  
In a similar vein for fairness metrics, we use one metric that does not consider the types of incorrect predictions (demographic parity, DP) and one that does take such information into account (false negative subgroup fairness, SFN).

We evaluated each of the three weighting methods on 11 datasets (including two medical datasets).
The collection of three reweighting methods, four combinations of evaluation metrics, and 11 datasets yields 132 experiments.  
We conducted 20 replicates for each of these experiments. Each replicate corresponds to a unique test-train split and includes 1000 model evaluations.  
We restricted our work to a predetermined configuration of a Random Forest Classifier model \cite{sup_material}, where different seeds can lead to unique forest structures for a given data split.

Parameters specific to each reweighting method are give as follows:
\begin{enumerate}
    \item \textbf{Equal Weights (EQ):} All data points are weighted equally (with a weight of 1.0). 
    For each of the 20 replicates, the model was evaluated $1000$ times using unique seeds.
    \item \textbf{Deterministic Weights (DW):} Sample weights are calculated using the procedure described in Section \ref{sec:reweigh}. 
    For each of the 20 replicates, the model was evaluated $1000$ times using unique seeds.
    \item \textbf{Evolved Weights (EW):} A population of 20 individuals (sets of sample weights) is evolved for 50 generations, resulting in $1000$ total model evaluations with our GA approach (Algorithm \ref{alg:rga}). 
    For each of the 20 replicates, the same model (i.e. all models use the same seed in a replicate) was run $1000$ times with different sample weights.
\end{enumerate}

Additionally, for a given dataset and replicate, each reweighting method is evaluated using models initialized with different random seeds. This design choice ensures that comparisons across methods are not confounded by identical random initializations.
Concretely, for dataset 
$i$ and replicate $j$, the Equal Weights (EQ) method uses models initialized with seeds $(i+j)*1000$,$(i+j)*1000+1$, and so on.
The Deterministic Weights (DW) method uses models initialized with seeds $(i+j+1)*1000$,$(i+j+1)*1000+1$, and so on.
The Evolved Weights (EW) method uses a model initialized with seed $(i+j+2)*1000$, with additional stochasticity arising from different sample weights evaluated across generations.

Our analysis in this paper builds on prior work with evolved weights~\cite{la2023optimizing,hoitsma2025mitigating} by analyzing the three weighting mechanisms under multiple combinations of predictive and fairness metrics, using Pareto fronts to assess the trade-offs in the error-fairness space.



\subsection{Real World Medical Datasets}

We use medical data on perinatal mood and anxiety disorders (PMADs) from \citet{wong2024} to validate the effectiveness of our method.
This data was collected through the Postpartum Depression Screening, Education, and Referral Quality Improvement Initiative at Cedars-Sinai Medical Center (CSMC) in Los Angeles, California, between 2020 and 2023.
We include data from birthing individuals who were admitted to the postpartum unit or the maternal-fetal care unit after delivery; individuals from the prenatal/pre-delivery time point or those who experienced stillbirth were not included.
The CSMC’s Institutional Review Board approved the use of de-identified patient data.
For our experiments, we use the two datasets corresponding to two methods of screening for PMAD. 
Below, we describe both screening questionnaires where higher scores indicate greater severity of depression. 

\textbf{\textit{Patient Health Questionnaire (PHQ-9)}}: The PHQ-9 consists of nine questions on a 4-point Likert scale (i.e., ‘not at all,’ ‘several days,’ ‘more than half the days,’ and ‘nearly every day’).
Aggregated scores range between $[0,27]$.

\textbf{\textit{Edinburgh Postnatal Depression Scale (EPDS-10)}}: 
The EPDS consists of ten questions on a 4-point Likert scale regarding the frequency with which respondents experienced symptoms of depression (e.g., `I have blamed myself unnecessarily when things go wrong').
Aggregated scores range between $[0,30]$.

\textbf{\textit{Target Variable}}:
Scores from the PHQ-9 and the EPDS were dichotomized into `low risk' (i.e., negative) or `moderate to high risk' (i.e., positive) according to each scale’s scoring criteria. 
Screening positive for depression risk was determined by endorsement of at least one of the following conditions being true: (1) suicidal ideation, (2) PHQ-9 $\geq 5$, or (3) EPDS $\geq 8$.

As recommended in \citet{wong2024}, to address the issues related to class imbalance, random undersampling was performed on the training set; the validation and test sets were not modified. 

\subsection{Publicly Available Benchmark Datasets}

\begin{table}[]
\caption{Set of publicly available datasets used in this study.}
{\begin{tabular}{lrrlrll}
\toprule
name & \#rows & \#cols & target name & favorable label & sensitive attributes\\
\midrule
heart\_disease       & 303                  & 13                    & target           & 1                                     & \{age \}            &  \\
student\_math      & 395                  & 32                    & g3\_ge\_10       & 1                                     & \{sex, age\}         &  \\
student\_por        & 649                  & 32                     & g3\_ge\_10       & 1                                     & \{sex, age\}         &  \\
creditg          & 1,000                & 20                     & class            & good             & \{personal...*, age\}         &  \\
titanic         & 1,309                & 13                   & survived         & 1                                     & \{sex\}          &  \\
us\_crime        & 1,994                & 102                  & crimegt70pct     & 0                                     & \{blackgt6pct\}        &  \\
compas\_violent & 4,020                & 51                    & two\_year\_recid & 0                                     & \{sex, race\}        &  \\
nlsy             & 4,908                & 15                   & income96gt17     & 1                                     & \{age, gender\}      &  \\
compas           & 6,172                & 51                & two\_year\_recid & 0                                     & \{sex, race\}        &  \\
\bottomrule
\end{tabular}}
\label{tab:datasets}
\end{table}

In addition to the medical datasets, in this study, we used a set of other open-access datasets for fairness evaluation from Hirzel \& Feffer~\cite{hirzel2023suite}.
Here, we focus on the datasets with binary targets and select the top 9 datasets when ordered by row count in ascending order.\footnote{We skip the \textit{ricci} dataset as most models produce a training AUROC score of 1.0 with it.}. 
Table \ref{tab:datasets} describes these datasets. 
Note that because there were no issues with class imbalance in these datasets, random undersampling was not performed on them.

\subsection{Hypervolume}
\label{sec:hypervolume}
\begin{figure}[!h]
\centering
\includegraphics[width=0.35\textwidth]{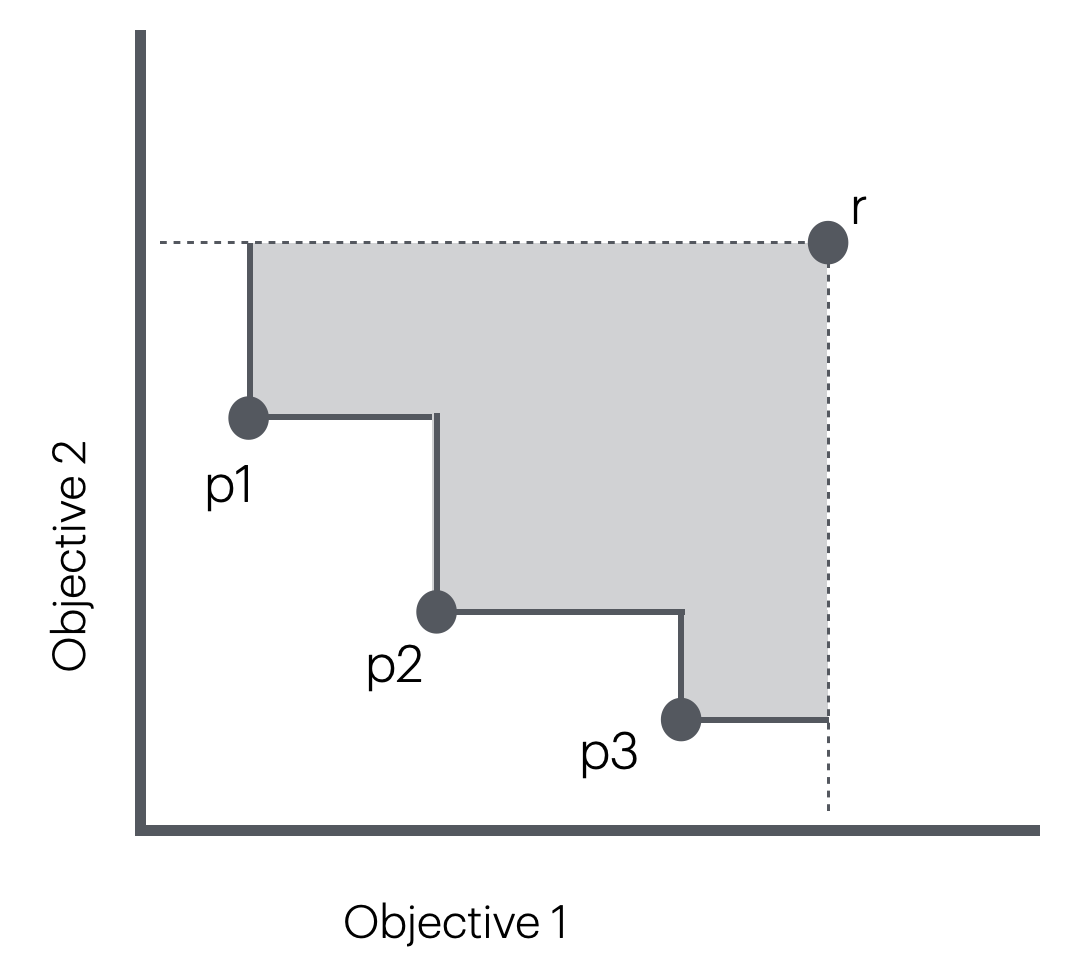}
\caption{Hypervolume for a given Pareto front comprising three points: $p1$, $p2$, and $p3$. The reference point ($r$) is used to calculate the hypervolume (shaded region) of the front.}
\label{fig:example_hv}
\end{figure}

We use hypervolume~\cite{fonseca2006improved} to assess the quality of the Pareto fronts generated by each weighting method.
Hypervolume summarizes each Pareto front into a single value by calculating the area of the objective space covered by each solution's performance within the front relative to a reference point.
Figure \ref{fig:example_hv} visualizes the hypervolume calculation for an example Pareto front optimized to minimize two objectives; the shaded area represents the hypervolume.

Pareto fronts with larger hypervolume are better for the following reasons. (a) Diversity~\cite{guerreiro2020hypervolume}: A larger hypervolume indicates that the Pareto front contains solutions that cover a wider range of trade-offs between the objectives. This diversity offers more options for decision-makers.
(b) Proximity~\cite{guerreiro2020hypervolume}: A larger hypervolume often indicates that the Pareto front is closer to the true optimal front. This means that the solutions are both diverse and closer to optima.
(c) Pareto-compliant~\cite{cao2015using}: Whenever the points on a Pareto front dominates the points on another Pareto front, the hypervolume of the former is more than that of the latter.

The procedure we use to compute the hypervolume is similar to the one described in La Cava~\cite{la2023optimizing}.
For every dataset and every experimental condition combination, we launch 20 replicates as described in Section \ref{sec:exp_conds}. 
Then, for each replicate, we consider the Pareto front obtained from all the evaluated models in that run using their values on predictive and fairness metrics on the test set, and calculate the hypervolume of that Pareto front. 
To ensure consistency in hypervolume computation, both metrics are converted into minimization objectives: $(1-predictive\_metric)$ and $fairness\_metric$.
A reference point of (1.0,1.0) is used for hypervolume calculation.

\subsection{Statistical analysis}
\label{sec:stats}
We conduct two types of statistical analyses in this study, as outlined in this section.
The first analysis employs non-hierarchical statistical tests to compare differences in the hypervolume of the Pareto fronts across weighting methods for a given dataset and a specific combination of predictive and fairness metrics.
The second analysis uses hierarchical modeling to assess overall trends, examining the effects of weighting methods, predictive and fairness metrics, and their interactions on hypervolume.

\subsubsection{Non-hierarchical Tests}
We conducted the Friedman test to detect significant differences in the hypervolume of the Pareto front between the weighting methods for a given dataset under a particular combination of metrics. 
If the Friedman test reported significant differences, we then performed a post-hoc paired Wilcoxon signed-rank test with a Bonferroni correction for multiple comparisons to identify differences among specific weighting method pairs. 
A significance level of 0.05 was used for all statistical tests.

\subsubsection{Hierarchical Tests}
\label{sec:mixed_effects}
\begin{figure}[!h]
\centering
\includegraphics[width=0.4\textwidth]{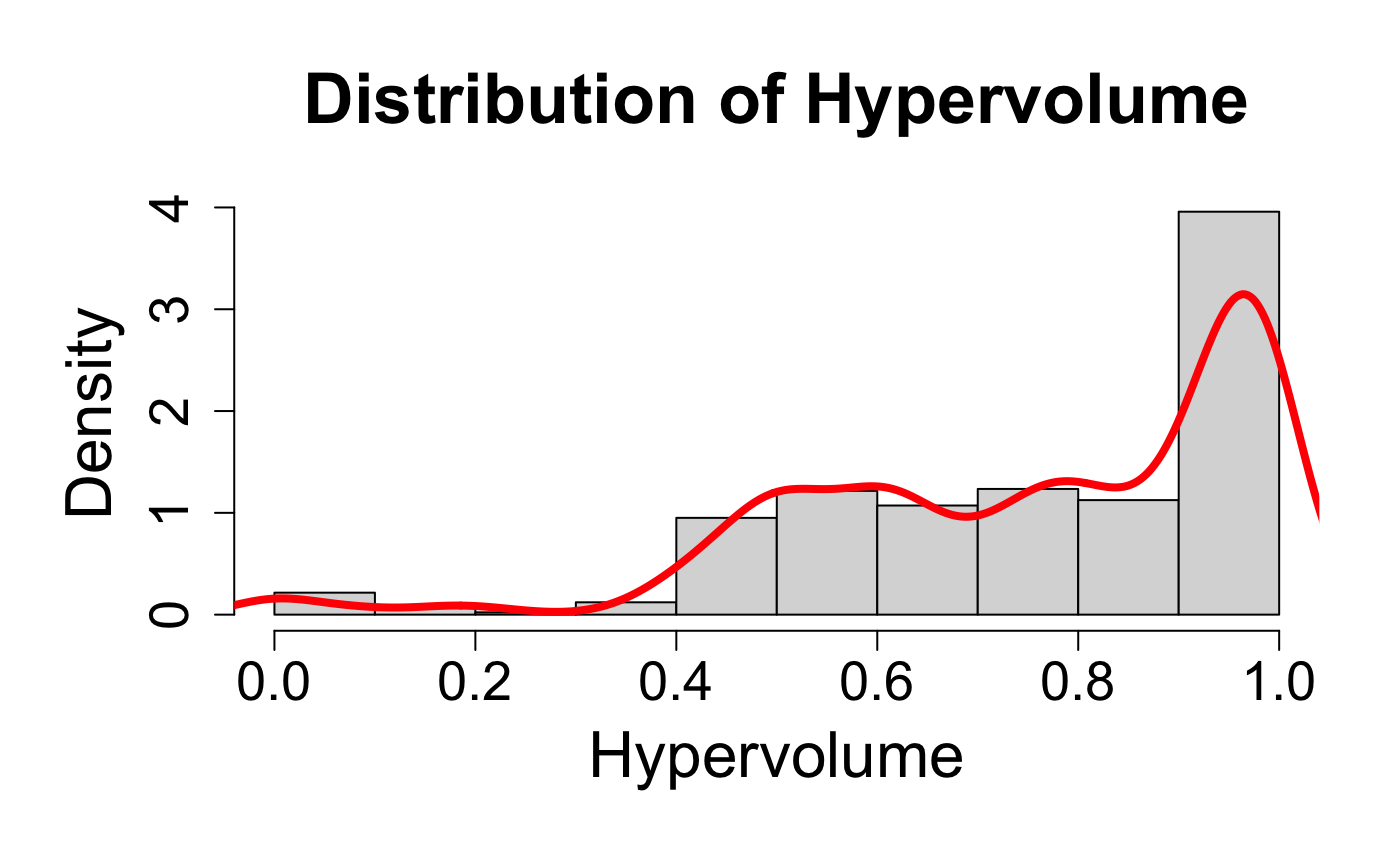}
\caption{Distribution of hypervolumes across all experimental runs.}
\label{fig:dist_hypervolume}
\end{figure}

To account for the hierarchical structure of our experimental results, we employed a Bayesian mixed-effects modeling approach based on a Zero-One Inflated Beta (ZIOB) regression framework. This choice is well-suited to hypervolume values, which are bounded between 0 and 1 and exhibit inflation near 1 (See Fig. \ref{fig:dist_hypervolume}).
In this model, hypervolume (\textit{hv}) is treated as the outcome variable, with the primary predictors of interest being the weighting strategy (with Equal Weights as the reference condition), the predictive objective (AUROC vs.\ Accuracy), and the fairness objective (subgroup false negative fairness vs.\ Demographic Parity). Formally, we model this relationship as
\[
hv \sim exp * objective1 * objective2 + (1 \mid dataset/rep),
\]
where \textit{exp} denotes the weighting method, \textit{objective1} denotes the predictive metric, and \textit{objective2} denotes the fairness metric used during optimization. The terms \textit{dataset} and \textit{rep} refer to the dataset and replicate index, respectively, such that \textit{rep} is nested within \textit{dataset}.
The fixed-effects component, $exp * objective1 * objective2$, includes all main effects and interaction terms, allowing us to test the influence of weighting strategies on hypervolume to vary across combinations of predictive and fairness objectives. The \textit{p}-values were computed according to Sivula et al. \cite{sivula2025uncertainty}. The random-effects component, $(1 \mid dataset/rep)$, captures structured variation by allowing baseline hypervolume levels to differ across datasets and across replicates nested within each dataset. 
This hierarchical structure accounts for correlations among repeated runs and prevents dataset- or replicate-specific variability from being incorrectly attributed to the experimental factors of interest.


\subsection{Software availability}


Our supplemental material is hosted on GitHub\footnote{https://github.com/theaksaini/Comparing-Reweighting-Methods} and contains all files related to software, data analysis, figure visualization, and documentation for this work.
\section{Results}
\label{sec:results}
By using the statistical tests in Section \ref{sec:stats}, for each dataset under consideration, we determine whether the evolved weights perform significantly better than other reweighting methods in terms of the hypervolume of the Pareto fronts (see Section \ref{sec:hypervolume}) for different combinations of metrics.
In Table \ref{tab:pvalues}, for each dataset that yielded significant differences according to the Friedman test, we present the results of the pairwise comparison between evolved weights and other methods using the Wilcoxon signed-rank test (the exact p-values are given in the supplementary material).

We denote the experimental conditions as follows. 
EW, DW, and EQ refer to the Evolved Weights, Deterministic Weights, and Equal Weights reweighting methods, respectively. 
ACC and ROC represent the predictive metrics accuracy and area under the ROC curve (AUROC), while DP and SFN correspond to the fairness metrics demographic parity and subgroup false negative fairness, respectively.
Table \ref{tab:pvalues} contains columns corresponding to the combinations of predictive and fairness metrics used as experimental conditions. 
For example, (ACC, DP) denotes the condition combining accuracy and demographic parity. These results are discussed in detail in the next section.

\begin{table}[]
\caption{Statistical significance of performance differences across datasets (EW: Evolved Weights, DT: Deterministic Weights, EQ: Equal Weights). EW>DT and EW>EQ indicate, respectively, that evolved weights yield significantly higher hypervolume than deterministic and equal weights (Wilcoxon rank-sum). Cells are populated only when the Friedman test detects significant differences (Sec.~\ref{sec:stats}).}
{\begin{tabular}{lllll}
\toprule
Dataset & (ACC, DP) & (ACC, SFN) & (ROC, DP) & (ROC, SFN)\\
\midrule
heart\_disease & EW>DT, EW>EQ &   & EW>DT, EW>EQ &EW>DT, EW>EQ \\
student\_math      & EW>DT, EW>EQ  &  &  EW>DT, EW>EQ  &  EW>DT, EW>EQ\\
student\_por        & EW>DT, EW>EQ &  &  EW>DT, EW>EQ& \\
creditg          & EW>DT, EW>EQ &  &  EW>DT, EW>EQ &   \\
titanic         & EW>DT, EW>EQ  & EW>DT &  EW>DT, EW>EQ &  EW>DT, EW>EQ\\
us\_crime        &EW>DT, EW>EQ &   &    EW>DT, EW>EQ  &   \\
compas\_violent & EW>DT, EW>EQ &  &  EW>DT, EW>EQ &       \\
nlsy             & EW>EQ &  &  EW>EQ&   EW>DT, EW>EQ \\
compas           & EW>DT, EW>EQ   &  & &  EW>DT, EW>EQ \\
pmad\_phq & EW>DT, EW>EQ  & EW>DT, EW>EQ  &EW>EQ &  EW>DT \\
pmad\_epds & EW>DT, EW>EQ  & EW>DT, EW>EQ  &EW>EQ &   \\
\bottomrule
\end{tabular}}
\label{tab:pvalues}
\end{table}

\begin{figure}[!h]
\centering
\includegraphics[width=1.00\textwidth]{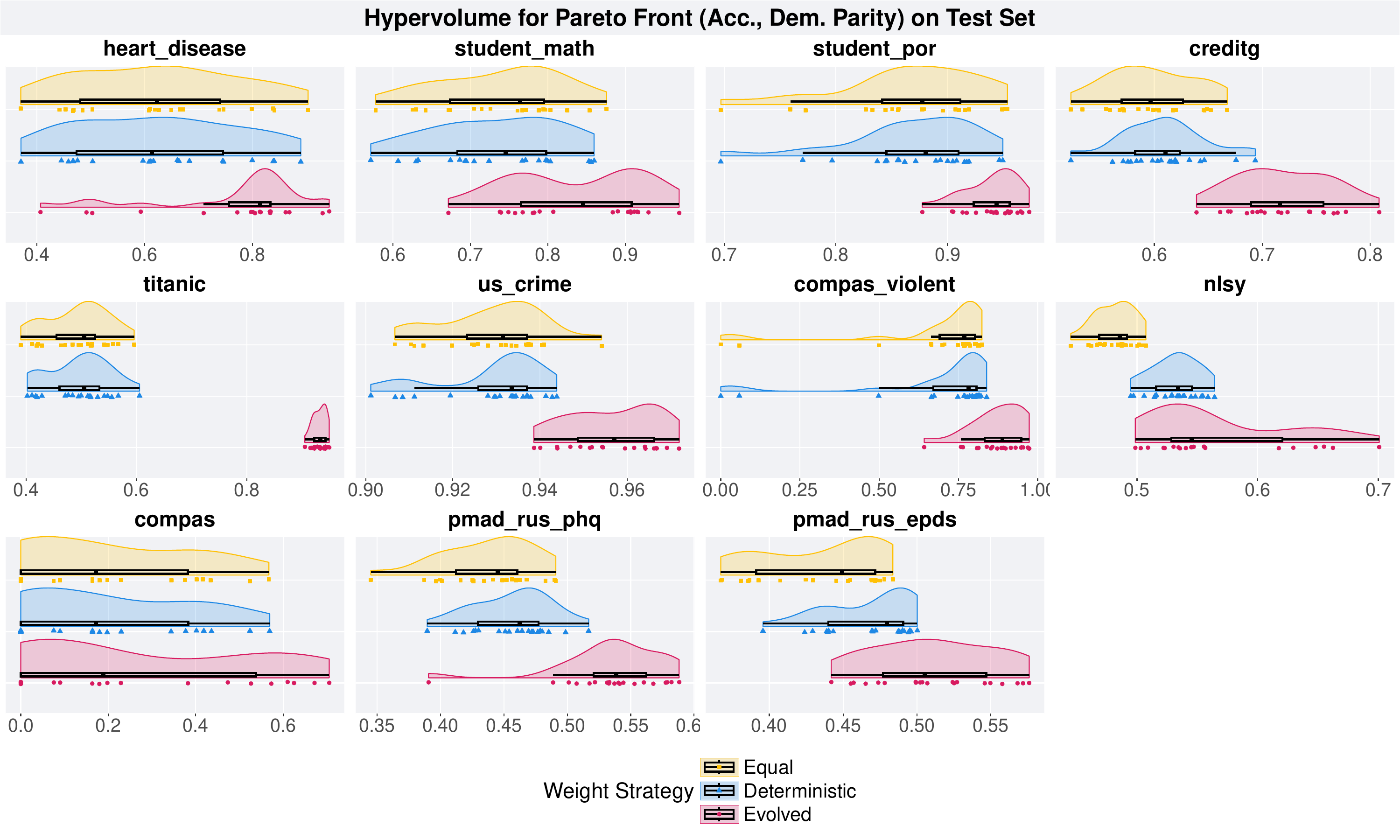}
\caption{Hypervolume for the Pareto fronts constructed from the performance on the test data for various datasets. The figure shows the experimental condition where accuracy and demographic parity have been used as the predictive and fairness metrics, respectively. Each point in the figure represents a single replicate.}
\label{fig:acc_dpd}
\end{figure}

Additionally, since the (ACC, DP) experimental condition leads to the maximum number of datasets where evolved weights performed significantly better than other methods, we also plot the hypervolume values for all datasets and weighting methods for this condition in Figure \ref{fig:acc_dpd}.
Each plot in the figure denotes the hypervolume of the Pareto front constructed from all the evaluated models in the corresponding dataset and weighting method. 

Moreover, as described in Sec. \ref{sec:mixed_effects}, to determine which experimental factors influenced test-set hypervolume, we fit a statistical model that relates hypervolume to the weighting method, the predictive objective, the fairness objective, and their interactions.
After fitting the model, we examined the estimated effects of each factor by summarizing how much the model output changed when that factor varied, while accounting for uncertainty in the estimates.
In other words, after a full model was fit that included all main effects and interactions, for each factor (or interaction), a reduced model was considered in which that specific term was removed.
The comparison between the full and reduced models quantified how much predictive performance changed when that factor was excluded.
This change is summarized in Table \ref{tab:mixed_effects} using a metric known as Expected Log Predictive Density Difference, or \textit{elpd\_diff} for short.
This table shows `Test of Effects,' with one row per factor or interaction. 
It summarizes which aspects of the experimental setup had a clear influence on hypervolume and which did not, providing an interpretable overview of the main drivers of performance differences across weighting strategies and objective choices.

\textbf{Omnibus Results}. In Table \ref{tab:mixed_effects}, we observe significant main effects of \textit{exp} (weighting method), \textit{obj1} (choice of predictive objective), and \textit{obj2} (choice of fairness metric). 
However, the effect of \textit{exp} (weighting method) is significantly moderated by \textit{obj2} (choice of fairness metric), and vice versa. The interaction between \textit{exp} and \textit{obj1} is not significant, nor is the interaction between \textit{obj1} and \textit{obj2}. Lastly, the three-way interaction between \textit{exp}, \textit{obj1}, and \textit{obj2} is not significant. 

\textbf{Probing Interaction Effects.} To better understand the interaction effect between \textit{exp} and \textit{obj2}, we conducted post-hoc pairwise comparisons with a Tukey correction for multiple comparisons. Table \ref{tab:two_way_probe} displays the pairwise comparisons between each weighting method across each fairness objective. For both SFN and DP, deterministic weights did not yield significantly different hypervolumes than equal weights. Evolved weights improved hypervolume relative to both equal and deterministic weights, with benefits significantly larger when the choice of fairness objective was DP.

\begin{table}[h]
\caption{Contribution of each experimental factor to differences in hypervolume. For each effect, \textit{elpd\_diff} reports the change in predictive performance relative to the reference model, \textit{se\_diff} indicates the uncertainty of this estimate, and \textit{z} represents the standardized effect size computed as the ratio of \textit{elpd\_diff} to \textit{se\_diff}. The column \textit{p} reports the corresponding two-sided probability associated with \textit{z}, denoting the strength of evidence. Larger absolute values of \textit{elpd\_diff} and \textit{z}, and smaller values of \textit{p}, indicate stronger evidence that the corresponding factor influences hypervolume.}
{\begin{tabular}{lllll}
\toprule
factor & elpd\_diff & se\_diff & $z$ & $p$\\
\midrule
exp	& -143.9703192 & 18.766223 & 7.671779 & 1.696269e-14\\
obj1	&-22.8675099 & 7.168822 & 3.189856 & 1.423437e-03\\
obj2	& -740.7891790 & 28.385510 & 26.097441 & 3.897947e-150\\
expXobj1	&-2.0940007 & 2.046388 & 1.023267 & 3.061817e-01\\
expXobj2&	-39.8681181 & 9.740546 & 4.093006 &  4.258159e-05\\
obj1Xobj2	&-3.9715006 & 3.905747 & 1.016835 & 3.092318e-01\\
3way	&-0.7338148 & 2.067999 & 0.354843 & 7.227072e-01\\
\bottomrule
\end{tabular}}
\label{tab:mixed_effects}
\end{table}

\begin{table}[]
\caption{Probing the effect of the weighting method across fairness objectives. The point estimates displayed are the median differences in hypervolumes. The final column represents the 95\% credible intervals; intervals that include 0 suggest no significant effect.}
{\begin{tabular}{lll}
\toprule
\textbf{objective2 = SFN:} & &  \\
\textit{contrast} & \textit{estimate} & \textit{95\% CI} \\
Equal Weights - Deterministic Weights & 0.00651  &[-0.094,    0.1026] \\
Equal Weights - Evolved Weights & -0.12514 &   [-0.226,   -0.0210] \\
Deterministic Weights - Evolved Weight &-0.12992&    [-0.238,   -0.0321] \\
\midrule
\textbf{objective2 = DP:} & &  \\
\textit{contrast} & \textit{estimate} & \textit{95\% CI} \\
Equal Weights - Deterministic Weights & -0.04518 &   [-0.136   , 0.0431] \\
Equal Weights - Evolved Weights & -0.69811&    [-0.785,  -0.6036] \\
Deterministic Weights - Evolved Weight & -0.64969 &   [-0.739,   -0.5598] \\
\bottomrule
\end{tabular}}
\label{tab:two_way_probe}
\end{table}
\section{Discussion}
\subsection{Evolution facilitates greater hypervolume}
The results given in Table \ref{tab:pvalues} demonstrate that for each of the datasets studied in this work, under at least one experimental condition (metrics combination), evolved weights lead to better optimization of predictive and fairness metrics than other reweighting methods.
In other words, the hypervolume of Pareto fronts obtained from models trained with evolved sample weights is higher than the hypervolume generated by alternative weighting methods for most of the combinations of dataset and experimental conditions.
The dependence between the hypervolume and the weighting method is also borne out in mixed-effects modeling described in previous sections. 
Specifically, Table~\ref{tab:mixed_effects} shows that the choice of weighting method (\textit{exp}) has a significant impact on the hypervolume.

Our results are consistent with other works where sample weights evolved using evolutionary algorithms have been shown to perform better than other fairness-enhancing methods, such as game-theoretic intervention GerryFair~\cite{kearns2018preventing}, in optimizing fairness and predictive performance~\cite{la2023optimizing}.

The success of the evolved weights approach is likely due to the more sophisticated search for optimal sample weights through the evolutionary mechanism, relative to deterministic weights calculated based on dataset characteristics. 
The GA presented here is designed to simultaneously optimize sample weights for both predictive performance and fairness, thereby optimizing a Pareto front. 
Additionally, the GA selects parents based on both predictive performance and fairness for a
particular ML model, which results in evolved sample weights tailored for that model; alternative weighting methods are model-agnostic. 
As a result, these compounding effects contribute
to the construction of Pareto fronts with greater hypervolume with the evolved weights approach compared to other methods.

One limitation of the genetic algorithm approach is that it requires multiple training calls to the model to evaluate the evolved sample weights, whereas other methods do not require as many calls. 
However, in circumstances where algorithmic fairness is paramount---such as healthcare or criminal justice---the reduction in algorithmic bias may justify the additional computational resources.

\subsection{Benefits of Evolved Weights are influenced by the Fairness Objective}
As shown in Table~\ref{tab:pvalues}, the number of datasets for which Evolved Weights (EW) outperform the two baseline methods depends on the choice of optimization objectives (metrics).
Specifically, EW yields significantly better results than other methods on 10 datasets under the (ACC, DP) objective pair, 2 datasets under (ACC, SFN), 7 datasets under (ROC, DP), and 5 datasets under (ROC, SFN).
This pattern suggests that the effectiveness of evolving sample weights is sensitive to the predictive–fairness metric combination.
Mixed-effects modeling further clarifies this relationship. As detailed in Section~\ref{sec:results} and Table~\ref{tab:mixed_effects}, while both objectives influence hypervolume, interaction effects reveal that the weighting method's impact is significantly modulated only by the choice of fairness objective.


This disparity in the performance of evolved weights across fairness metrics likely stems from the specific type of fairness they measure and their interactions with predictive metrics. 
Some fairness metrics permit fairness gains without significantly compromising predictive performance, thereby expanding the Pareto front and hypervolume.
This is especially evident for DP, which compare predictions \textit{within} subgroups (e.g., acceptance rate) rather than \textit{between} subgroups and the overall model performance.
As a result, this metric can sometimes be improved not by making the model more accurate or equitable, but by uniformly worsening the predicted outcomes for all subgroups, without affecting overall predictive performance. 
This creates the appearance of fairness improvement even when subgroup-level predictions degrade.

\begin{table}[]
\caption{Toy Example}
{\begin{tabular}{ccccc}
\toprule
\multirow{2}{*}{Group} & \multicolumn{2}{c}{Before Reweighting} &
      \multicolumn{2}{c}{After Reweighting} \\
&$y$  &$\hat{y}$&$y$ &$\hat{y}$ \\
\midrule
A & 1 & 0 & 1 & 0  \\
A      & 0  & 0 &  0 & 0 \\
B  & 0  & 0 &  0 & 0 \\
B         & 0 & 1 &  0& 0  \\
B         & 1  & 1 & 1 & 0 \\
B        &1 & 1  & 1    &   1\\
\bottomrule
\end{tabular}}
\label{tab:toy_example}
\end{table}

The toy example in Figure~\ref{tab:toy_example} illustrates this effect.
Consider a dataset with six instances from two groups, $A$ and $B$, with true labels $y$ and predicted labels $\hat{y}$.
Before reweighting, the model's accuracy is 0.67. The acceptance rates are 0 for group $A$ and 0.75 for group $B$, giving a DP of 0.75.
After reweighting (using GA), the model's accuracy remains 0.67, but group $B$’s acceptance rate falls to 0.25 while group $A$’s remains at 0. This reduces the DP to 0.25.
From a metric standpoint, fairness has improved; however, the actual outcomes for both groups have become strictly worse.
This highlights a broader issue: some fairness metrics can be improved through degenerate strategies that do not genuinely benefit the affected groups.

These observations suggest that the choice of optimization objectives plays a crucial role not only in how easily fairness can be improved, but also in what kind of improvements are encouraged by the metric definitions themselves.
Understanding these interactions is essential when evaluating fairness-enhancing interventions such as reweighting.

Future work should investigate whether improvements in fairness metrics achieved through various reweighting strategies translate into genuinely improved outcomes for all subgroups.
In addition to the objectives used during optimization, the performance of evolved weights may also depend on the particular parameter settings used in the GA configuration.
Accordingly, future investigation could replicate our experiments using different ML models and explore alternative GA configurations, including adjustments to parameters such as the parent-selection strategy (e.g., replacing NSGA-II with lexicase selection~\cite{helmuth2014solving}).

\section{Conclusions}
In this paper, we compared three approaches for assigning sample weights during the training of machine learning models---a process known as reweighting---which is a commonly used fairness enhancement technique.
Using eleven publicly available datasets (including two from the medical domain), we evaluated a Genetic Algorithm approach for evolving sample weights, termed Evolved Weights (EW), against two baseline strategies: Equal Weights (EQ), which assigns equal weights to all samples, and Deterministic Weights (DW), which computes weights solely from data characteristics.
We evaluated these three reweighting strategies under four combinations of predictive and fairness metrics, using accuracy and area under the ROC curve (AUROC) as predictive metrics, and demographic parity and subgroup false negative fairness as fairness metrics. 
Pareto fronts were computed for each dataset in the corresponding predictive–fairness objective space.
Our experiments demonstrate that EW leads to more effective optimization of the trade-offs between fairness and predictive performance.
Specifically, for every dataset, EW produced significantly better Pareto fronts than other techniques under at least one combination of metrics.
However, the magnitude of these improvements, indicated by the number of datasets for which Evolved Weights (EW) outperform the two methods, depends on the specific objectives (metrics) used during optimization.  
For further analysis, we also performed mixed-effects modeling based on  Zero-One Inflated Beta (ZIOB) regression framework and observed significant main effects of the weighting method, the choice of predictive objective, and the choice of fairness objective. 
Additionally, the analysis revealed that the effect of the weighting method was significantly moderated by the choice of the fairness metric.
Overall, these findings suggest that evolutionary optimization of sample weights can enhance fairness without substantial loss in predictive accuracy, while also highlighting the critical role of metric selection in determining fairness and predictive performance outcomes. 
Future work should investigate additional metric combinations, analyze the mechanisms underlying metric-dependent improvements, and explore the generalizability of the EW approach across different model architectures and application domains.


\clearpage
\bibliographystyle{ACM-Reference-Format}
\bibliography{references}

\appendix
\section{Probing additional interaction effects}
Table \ref{tab:three_way_probe} describes the effect of the weighting method across different combinations of performance and fairness objectives.
\begin{table}[h]
\caption{Probing the effect of the weighting method across different combinations of performance and fairness objectives. The point estimates displayed are the median differences. The final column represents the 95\% credible intervals; intervals that include 0 suggest no significant effect.
\label{tab:three_way_probe}}
{\begin{tabular}{lll}
\toprule
\textbf{objective1 = AUROC, objective2 = SFN:} & &  \\
\textit{contrast} & \textit{estimate} & \textit{95\% CI} \\
Equal Weights - Deterministic Weights & 0.00932 &   [-0.121,    0.1624] \\
Equal Weights - Evolved Weights & -0.05259 &    [-0.193,    0.1023] \\
Deterministic Weights - Evolved Weight & -0.06134&    [-0.208,    0.0852]\\
\midrule
\textbf{objective1 = ACC, objective2 = SFN:} & &  \\
\textit{contrast} & \textit{estimate} & \textit{95\% CI} \\
Equal Weights - Deterministic Weights & 0.00204 &   [-0.147,    0.1418] \\
Equal Weights - Evolved Weights & -0.19728  &  [-0.343,   -0.0467] \\
Deterministic Weights - Evolved Weight & -0.19997 &   [-0.352,   -0.0548] \\
\midrule
\textbf{objective1 = AUROC, objective2 = DP:} & &  \\
\textit{contrast} & \textit{estimate} & \textit{95\% CI} \\
Equal Weights - Deterministic Weights & -0.04615   & [-0.173,    0.0836] \\
Equal Weights - Evolved Weights &  -0.74535   & [-0.875,   -0.6136] \\
Deterministic Weights - Evolved Weight & -0.70025   & [-0.818,   -0.5535] \\
\midrule
\textbf{objective1 = ACC, objective2 = DP:} & &  \\
\textit{contrast} & \textit{estimate} & \textit{95\% CI} \\
Equal Weights - Deterministic Weights & -0.04463   & [-0.169,    0.0800]\\
Equal Weights - Evolved Weights & -0.64860   & [-0.774,   -0.5261] \\
Deterministic Weights - Evolved Weight & -0.60240   & [-0.721,   -0.4749] \\
\bottomrule
\end{tabular}}
\end{table}

\end{document}